\documentclass[runningheads]{llncs}
\usepackage{pdfpages}
\usepackage{graphicx}
\graphicspath{ {images/} }
\usepackage{dsfont}
\usepackage{bm}
\usepackage[utf8]{inputenc} 
\usepackage[T1]{fontenc}    
\usepackage{hyperref}       
\usepackage{url}            
\usepackage{booktabs}       
\usepackage{amsfonts}       
\usepackage{nicefrac}       
\usepackage{microtype}      
\usepackage{amssymb}
\usepackage{amsmath}
\usepackage{multirow}
\usepackage[]{algorithm2e}
\usepackage{afterpage}
\usepackage{float}
\usepackage{authblk}

\usepackage[utf8]{inputenc}
\usepackage[T1]{fontenc}
\usepackage[english]{babel}
\usepackage{graphicx}
\usepackage{amsfonts,amsgen,amsmath,amssymb}
\usepackage{afterpage}
\usepackage{colortbl,longtable}
\usepackage{url}
\usepackage[stable]{footmisc}
\usepackage{parskip} 
\usepackage{hyperref}
\usepackage{nameref}
\usepackage{amsmath,amsfonts,bm} 
\usepackage{booktabs}
\usepackage{mathtools}

\usepackage{fancyhdr}

\begin{document}

\title{An artificial consciousness model and its relations with philosophy of mind}
\author{Eduardo C. Garrido-Merchan\inst{1}\and
Martin Molina\inst{2}\and Francisco M. Mendoza\inst{3}}
\date{December 2020}

\institute{Universidad Autonoma de Madrid, Madrid, Spain
\email{eduardo.garrido@uam.es} \and
Universidad Politecnica de Madrid, Madrid, Spain
\email{martin.molina@upm.es} \and
Universidad Politecnica de Madrid, Madrid, Spain
\email{franciscomanuel.mendoza.soto@alumnos.upm.es}}

\maketitle

\begin{abstract}
This work seeks to study the beneficial properties that an autonomous agent can obtain by implementing a cognitive architecture similar to the one of conscious beings. Along this document, a conscious model of autonomous agent based in a global workspace architecture is presented. We describe how this agent is viewed from different perspectives of philosophy of mind, being inspired by their ideas. The goal of this model is to create autonomous agents able to navigate within an environment composed of multiple independent magnitudes, adapting to its surroundings in order to find the best possible position in base of its inner preferences. The purpose of the model is to test the effectiveness of many cognitive mechanisms that are incorporated, such as an attention mechanism for magnitude selection, pos-session of inner feelings and preferences, usage of a memory system to storage beliefs and past experiences, and incorporating a global workspace which controls and integrates information processed by all the subsystem of the model. We show in a large experiment set how an autonomous agent can benefit from having a cognitive architecture such as the one described. 
\end{abstract}

\section{Introduction}
Artificial consciousness \cite{chella2013artificial} is an emerging multidisciplinary area that propose computer science implementations to models coming from neuroscience \cite{crick1998consciousness} and discussed in philosophy of mind \cite{chalmers2002philosophy}. With the recent advancement of technology and machine learning methods \cite{murphy2012machine}, the field can propose new implementations of cognitive architectures using the latest proposed methodologies and techniques. 

Not only has the artificial consciousness advances but also discussions coming from philosophy of mind regarding consciousness \cite{kim2018philosophy}. As artificial intelligence \cite{russell2002artificial} and neuroscience advance, new questions are being discussed by philosophers over these improvements. Nevertheless, at least as far as we know, this does not happen the other way round and computer science does rarely use philosophy of mind and neuroscience ideas to keep developing its methods. For example, deep learning methods do not worry about the hypothesis that if living beings are the result of evolution and are conscious, then consciousness must give an evolutionary advantage over mere unconscious proccesing of information such as the one happening in deep learning. 
The motivation of the present work is to explore the utility of artificial consciousness models that in the future could be used, for example, to improve the quality of autonomous robotic systems (e.g., better performance efficiency). We do not intend to deliver a state-of-the-art deep learning method but only to start exploring how computer science models may imitate the flow of information that living beings do to see whether that processing of information produces a reasonable behaviour. The obtained results from the experiments of the implementation of our approach may also be useful to provide quantitative evidence that help to justify the need of consciousness in biological beings. We also discuss how our proposed cognitive architecture relates with the main philosophy of mind ideas regarding consciousness in an attempt to being a bridge between the two disciplines and motivate the study and implementation of philosophy of the mind ideas in computer science.

Previous work has been done regarding artificial consciousness \cite{reggia2013rise,soto2017artificial}. The literature contains the implementation of robots simulating correlated behaviours of beings exhibiting consciousness \cite{seth2009explanatory}. Other works include artificial consciousness cognitive software architectures such as a neuronal model of a global workspace for cognitive tasks \cite{dehaene1998neuronal}. The consciousness prior model \cite{bengio2017consciousness} is an alternative to this approach based in sparse factor graphs that have some similar elements to our proposed architecture. It will be further discussed in the philosophy of mind section of this paper. Machine learning \cite{alpaydin2020introduction} and deep reinforcement learning \cite{li2017deep} techniques are able to manage a big quantity of information and could be incorporated in both our proposed architecture and in other ones. Artificial consciousness models are classified into 4 levels depending on the characteristics of the implementation. Cognitive architectures involving different modules that exhibit behaviour correlated with consciousness are classified as level 3 systems and system that can arise potential phenomenal experiences would be considered level 4 machine consciousness system \cite{gamez2018human}. Our proposed approach is a level 3 cognitive architecture system as, from our point of view, we do not have the tools to determine whether level 4 systems could be implemented in practice. Although, this will be discussed in the philosophy of mind section of this work. 

This manuscript is organized as follows: We have already discussed related work and an introduction to artificial consciousness. We will now propose the different involved modules and models of the cognitive architecture for a potentially conscious agent and possible enhancements of them. We then present the cognitive architecture that consist on the previously described modules. Once that the model has been explained, we discuss the connections that our proposed approach have with philosophy of the mind theories and how our model fits some of these theories, like the functionalism theory. We continue with an exhaustive set of experiments where we show how an agent implementing our architecture exhibits an autonomous behaviour. Finally, we conclude the paper with some conclusions and future work proposals.

\section{Modules of the cognitive architecture of the robot}

Before describing the modules of the robot, we need to define the environment where it is going to be placed. The environment is a finite \(2D\) space represented as a grid $\mathbf{G}$ of dimension \(N \times N\) cells. Each cell contains observable physical magnitudes. In a real world, for example, these magnitudes could correspond to temperature, humidity, light, noise, presence of obstacles, etc. We use abstract magnitudes identified by \(m_1, .., m_n\). The value of each physical magnitude is represented by a normalized real number in the interval \([-a, +a]\)(for example, \( a=10\)).

The robot is able to perceive magnitudes using sensors. When the robot is located in a particular cell, it perceives information of each adjacent cell from every direction \((up, down, left, right)\). If, for example, there are five magnitudes, the robot perceives a total of 20 values (4 directions multiplied by 5 observed magnitudes). The robot is able to move in the environment by doing one of four basic motion actions: \(up, down, left, right\) in a particular instant. We represent with $\mathcal{X}$ the set of possible actions that the robot may take. 

Having described the environment and how the robot moves, we describe the modules of our proposed implementation of a global workspace model architecture. These modules are an attention process that filters information that is not necessary for the robot to make a conscious decision, the long and short term memories, the sensors that the robot uses to detect information from the environment, the evaluation system, the motor system, the degree of happiness that the robot possesses and the most important process, the global workspace process.

We define that a robot has a set of feelings. In this model, we only consider one type of feeling that corresponds to the happiness degree. It is represented as $h \in \mathbb{R} : h \in [0,10]$ and can be initialized, for example in $h=5$ in such a way that higher values are good and lower are bad. If happiness degree decreases too much passing a threshold $\tau$, for example $\tau=2$, the robot deactivates, simulating death. As explained below, in this setting, the objective of the robot is to optimize its happiness degree $h$ w.r.t a single step in the grid $\mathbf{G}$. In other words, the decision $x^*$ that the robot takes follows $x^* = \arg \max_{x \in \mathcal{X}} f(x,t|h,\mathbf{\Theta}, \mathbf{p})$, where $\mathcal{X}$ is the set of possible actions that the robot may take in time $t$ conditioned to its happiness degree $h$ and in multiple flows of information $\mathbf{\Theta}$, which are the subconscious importance that the robot gives to every possible action $x$, and the short and long term memory about the actions that the robot is aware of. Lastly, $\mathbf{P}$ is a matrix of preferences that describes, for every sensor $s_k$ of the grid $\mathbf{G}$, the values of the sensor that make the objective $h$ of the robot change. For the shake of simplicity, in this implementation we have used a single objective $f(\cdot)$ (with the degree of happiness $h$) but in a more realistic situation this should be represented by a set of objectives $\mathbf{f}$. The problem that appears is that, for any time $t$, we can only evaluate $f$ for a single action $x$. As it is going to be explained later, this function $f$ is computed by the evaluative system and $f$ is estimated for $\mathcal{X}$ by the attentional system $a(\mathcal{X}|\mathbf{\theta}) \approx f$ where $\mathbf{\theta}$ is the set of weights of the attentional system and also estimated for the most relevant actions according to the attentional system by the global workspace model $g(a(\mathcal{X}|\mathbf{\theta})|\mathbf{\Theta}) \approx f$, employing all the information in $\mathbf{\Theta}$, that includes weights for the most relevant actions decided by the attentional system (subconscious decision) $a(\mathcal{X}|\mathbf{\theta})$, the short $\mathbf{s}(t)$ and long term memory $\mathbf{l}(t)$ and the current state of the objective that we are maximizing $h(t)$. The robot hence want to maximize, for every step $t$, the feeling $h$, emulating happiness, through $f(\cdot)$ that can only be executed once for every $t$ by the evaluative system. In order to make a proper decision to maximize $h(t)$, it maximizes $a(\cdot)$ and $g(\cdot)$ which are approximations to $f(\cdot)$ based in the cognitive information learned from the environment by the robot.

Each robot has its preferences $\mathbf{P}$ that represent how good a certain value given by a sensor $s_k$ observed in the environment grid $\mathbf{G}$ is for the robot. Each position $g_{ij}$ of the grid contains magnitudes observed by sensors. Preferences are functions of the magnitudes that affect how the situation is evaluated by the robot. The evaluative system conditions the evaluation of a position and magnitude $(g_{ij}, p(m_k))$ to the preferences of that sensor of the robot $\mathbf{p}_k$. Each robot can have different preferences. More details about the parametric expressions of these functions are given in the experiments section.
 
A belief is a conscious conclusion inferred by the robot that can be stored in the memory as experience to be used later as a tuple $(r|m_l,s_k,t)$, that is, a movement $m_l$ (up, down, left, right) exploring a sensor $s_k$ at time $t$ that has been evaluated with reward $r$. This conclusion is given by the evaluative system as a result of evaluating $f(x^*,t|h,\mathbf{\Theta}, \mathbf{p})$.  The information that we need to store and that conditions all the weights of the system is basically how good or bad is moving to a certain direction $m_l$. These beliefs tuples, $B$, are going to be stored in the long term memory $\mathbf{l}$ if a threshold $\beta_l$ is reached when we evaluate an action by the evaluative system $|f(m_l,s_k)| > \beta_l$ or in the short term memory with another threshold. Beliefs are ranked in the memory according to their relevance $f(\cdot)$. To simulate forgetting beliefs, we use the parameter \(l\) that represents the maximum number of beliefs in the memory. In the long term memory, beliefs that are less relevant are deleted first.  When beliefs have the same degree of relevance, oldest beliefs are deleted first. In the short term memory, new beliefs substitute older beliefs independently of their degree of relevance.

The reactive attention expresses how the robot pays attention \cite{american1998attention} to observations of sensors $s_k$ performed on each position of the grid $g_{ij}$. The attentional system represents the subconscious decision about the importance of all the inputs given by the perceptual system of the robot. The subconscious system of the robot pays attention to all the information of every position that the robot perform in time $t+1$ and of all the sensors that the robot can evaluate in time $t+1$. That is $m$ possible movements multiplied by $k$ sensors generating $mk$ possible actions that the robot can perform at time $t$. The consciousness bottleneck limits the amount of information that a human being can pay attention to in a certain time $t$, but a human brain is processing all the information given by the human senses, that is, the perceptual system of the robot. This is done in the subconscious mind, here the attentional system of the robot. This attentional system then is going to select $A$ possible actions from the available $mk$, for example $A=2$ that is going to transfer to the global workspace model, which is representing the awareness experience of the human being. In order to select which are the most important actions $x$, every possible $mk$ action is represented by a weight $w_{i} \in [0,1] : \sum_{i=0}^{mk} w_{i} = 1$, that is, a multinomial distribution of weights that is initialized randomly or uniformly. The decision about which is the most important action is done probabilistically according to that weights and to the inputs given by the perceptual system $x_{mk}$. The attentional system does not have access to the evaluation system because the action has not been performed and the robot has not taken a conscious decision about the next step.

We sample $A$ indexes from that multinomial distribution and that indexes determine the actions that are going to be transferred to the global workspace $a \sim M(\mathbf{w})$. These weights are also going to be conditioned by the happiness degree of the robot to represent the survival instinct of consciousness. If the robot has a value for $h$ far from the death threshold, then, exploration of new sensors and movements is encouraged and the selection is done probabilistically as described.  However, if the robot feeling value is close to death, $h < \tau + \epsilon$ then the selection will be done with a ranking of the weights. The idea is to exploit the actions that have been good in the past. 

Another mechanism that needs to be described of the attentional system is how are the weights $\mathbf{w}$ adapted from the evaluations done by the evaluation system $f(\cdot)$. These adaptations $\delta$ of the weights are going to be determined by the quality of the evaluation $f(\cdot)$ and the feeling objective value $h$, that is $\delta = e(f(\cdot),h)$. For $e(\cdot)$, we propose the following function, if the robot is not close to death $h \geq \tau + \epsilon$, then, we propose a learning rate sampled from a Gaussian distribution $r \sim N(0.05,0.001)$. If the evaluation done by the evaluative system $f$ has lower value than the short and long term memory thresholds and is positive $f(\cdot)>0$, the selected weight is incremented by $w_* = w_* + r$ and the rest of the weights are normalized $\mathbf{w}_n = \mathbf{w}_n - \frac{r}{mk-1}$ where $\mathbf{w}_n$ does not contain $w_*$. If it is negative $f(\cdot)<0$, the selected weight is decremented by $w_* = w_* - r$ and the rest of the weights are normalized $\mathbf{w}_n = \mathbf{w}_n + \frac{r}{mk-1}$ where $\mathbf{w}_n$ does not contain $w_*$.

If the evaluation done by the evaluative system has involved storing a belief in the short term memory $|f(\cdot)|>\beta_s$, then the weights are varied with a higher learning rate, for example: $r \sim N(0.1,0.001)$. If the evaluation is extreme, involving the long term memory, the learning rate is also higher, for example, $r \sim N(0.2,0.001)$.
Lastly, a special update process occurs when the robot is close to death, modifying drastically the weights in order to not continuing a bad behaviour or encouraging a good behaviour. In this scenario, the weights are modified by an extreme learning weight which is $min(1,N(1-(h-\tau),0.001))$.

The selected tuples $(m_i,s_k)$ fed to the global workspace are weighted for this system by a quantity $w_a \in [0,1] : \sum_{a=0}^{A} w_a = 1$ proportional to the sum of the weights of the selected tuples. More formally, the computation that needs to be done is: $w_a = \frac{w_a}{\sum_{a=0}^{A} w_a}$.

The goal of the robot is to move in the environment in order to optimize its feelings. In our case, it corresponds to maximize the degree of happiness $h$ as a single objective $f$. In a more realistic scenario, the robot would use a set of feelings $\mathbf{f}$ which correspond to a multi-objective optimization.  

At every time step $t$, the robot updates the happiness by the evaluation of a grid position and a sensor. The current state of happiness $h(t)$ is updated by the reward obtained by the evaluation system $r$ as $h(t+1) = h(t) + r$. In addition, at every time step $t$, the robot loses a fixed quantity of happiness, for example $c \sim N(0.01,0.001)$. The complete operation over the objective at every time $t$ is $h(t+1) = h(t) + r - c$. To avoid death, the robot must ensure that its happiness is above a configurable threshold $\tau$. That is the reason why, in order to survive, the robot must be always moving. 

The perceptual system gets information of the grid $g_{ij}$ and the sensors $s_k$ in the form of observations of physical magnitudes given by the sensors of the robot and fed to the attentional system (subconscious mind) and to the evaluation system.  

The global workspace makes a conscious decision about the action that is going to be performed at time $t$ with respect to the $A$ tuples $(m_i,s_k)$ given by the attentional system based on the importance given to those tuples by the attention system $a(\mathbf{x}|\mathbf{w})$, the short term memory $s(\mathcal{B}_s)$, where $\mathcal{B}_s$ is the set of short term beliefs, the long term memory $l(\mathcal{B}_l)$ and the degree of happiness $h$. The phenomenal experience (or subjective experience) is the following decision problem: which action to take \((up, down, left, right)\) based on the \(a\) tuples about observed physical magnitudes, or sensors $s_k$ and directions $m_i$ served by the attention module. We do only select $A$ tuples here in order to emulate the bottleneck of consciousness, although we are experiencing a bubble of data at each time $t$, we are only aware of a subset of this information, represented here by these tuples.

In order to make the decision of what is the tuple that I am most interested, that is, the action that I want to make at time $t+1$ and the action that needs to be ordered to do to the motor system, the global workspace assigns weights to every module involved in the decision, representing the importance that we consciously give to our instincts, represented by the attentional system $a(\cdot)$ and weighted as $w_a$; to the short term memory beliefs (if that tuple is contained in the set of short term memory beliefs) $w_s$ and to the long term memory $w_l$ these weights must sum $1$.

Every action $(m_l,s_k)_a$ will be assigned the following punctuation \\ $g(a(\cdot)|a(\mathbf{x}|\mathbf{w}), s(\mathcal{B}_s), s(\mathcal{B}_l), h(t), t) = f_h(w_a a(\cdot) + w_s s + w_l l|h)$, where $a(\cdot)$ is the importance given to the tuple by the attentional system, $s=1$ if the short term memory has determined that the tuple is positive and $s=-1$ if it is negative, $l=2$ if the long term memory has determined that the tuple is positive and $l=-2$ if it is negative. The function $f_h$ depends on the result of this operation and the degree $h$ of happiness. If the robot is in a good state, for example $h>5$, scores are not modified to encourage exploration. If it lies in a risk state $h<5$ and $h>\tau+\epsilon$, then the scores are modified by doubling the best score and decrementing to a half the worst score. If the situation is critical $h \in [\tau, \tau+\epsilon]$ the score given to the best tuple is $1$ and the other tuple recieves value $0$ just to focus on the best tuple.

The final decision is also done probabilistically, simulating rational doubts. Scores of the $A$ tuples are normalized with respect to its sum are recomputed as: $w_a = \frac{w_a}{\sum_{a=0}^{A} w_a}$. Then, we select at random from the multinomial distribution given by that weights.

At the beginning each module is initialized a weight uniformly. The global workspace then outputs the selected tuple to the perception system in order for it to inform the values of the sensor to the evaluation system and to the motor system to move to the new position $m_l$ and measure the sensor $s_k$.

The weights of every module $w_a,w_s,w_l$ are updated depending on the reward given by the evaluative system. If the criterion chosen by the attention was positive, the attention weight is incremented by $0.05$ and the others decremented and normalized as in the attention module. The contrary operation is done if the reward is negative. If stored beliefs corresponding to the selected tuple exist in the memories and they have the same criterion as the evaluative system, the memories are updated by the same mechanism as in the previous weight and viceversa. The update procedure is done iteratively and with a random order to not give preference to any particular module. This global workspace process corresponds to the System 2 module described in the Kahneman architecture \cite{kahneman2011thinking}. 

Finally, the goal of the evaluative system is to evaluate the current information fed by the perceptual system that the global workspace has selected $f(g(\cdot)| \mathbf{p})$, this evaluation modifies the degree of happiness, which is the feeling that is being optimized and it can generate beliefs that can be stored in the short-term and long-term memories. Once an action is decided to be taken by the global workspace, the observed physical magnitude that has been used in this decision is sent to the evaluative system to be potentially remembered as experience if the evaluation $f(\cdot) > \beta_s$ and $f(\cdot) > \beta_l$ for the short and long term respectively. Also, depending on $f(\cdot)$ the weights of the attentional system and the global workspace are modified as it has been described in their respective sections of this manuscript. 

The operation performed by the evaluative system is the following one. First, we apply the preference function that corresponds to the sensor $s_k$ that the perceptual system has measured, that is $r = \mathbf{p}_k(s_k(g_{ij}|t))$, obtaining the reward $r$ that modifies the happiness degree.
\section{Cognitive architecture}
\begin{figure}[htb!]
\begin{center}
\includegraphics[width=0.75\textwidth]{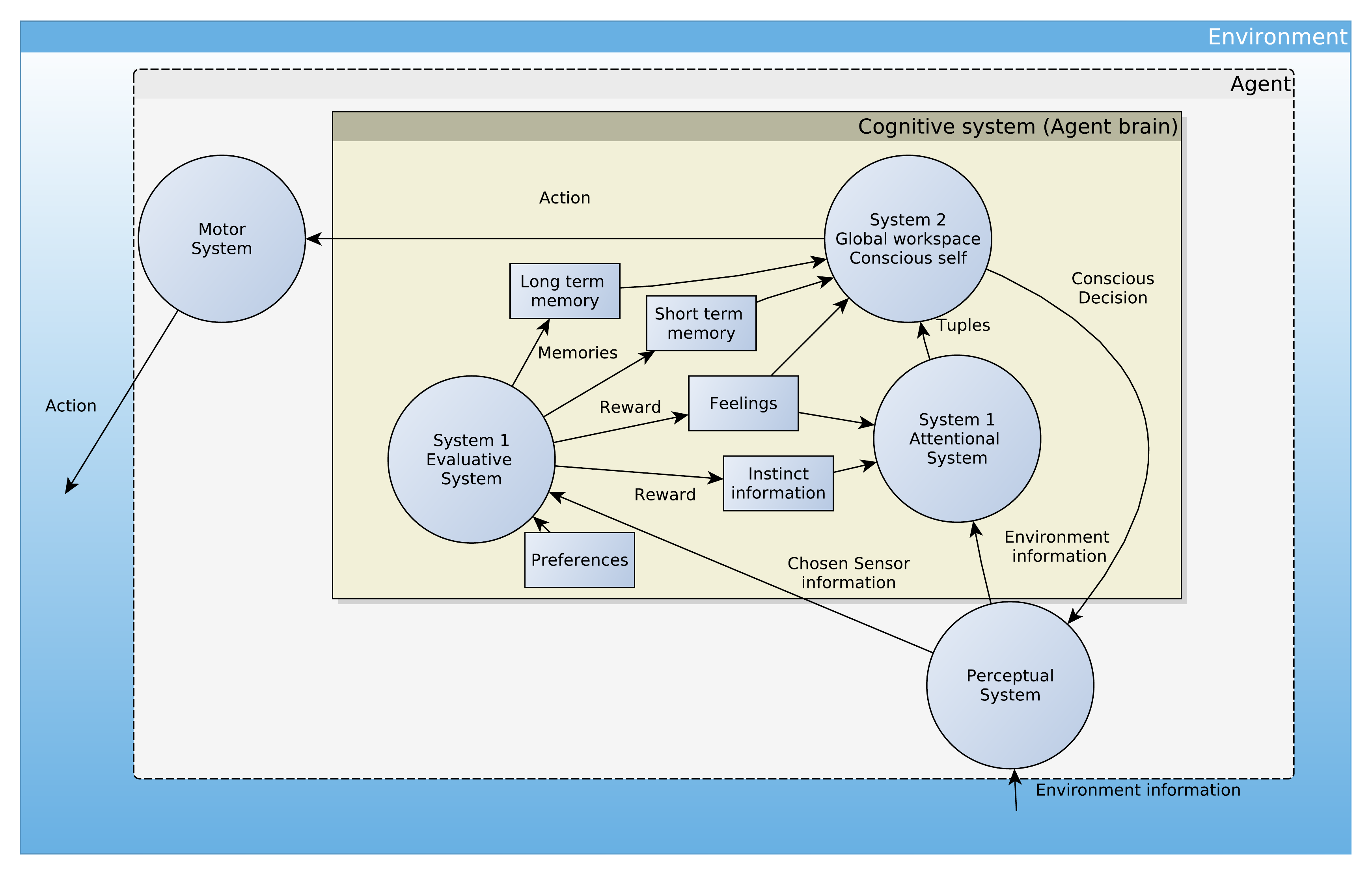}
\caption{Architecture of the models described in the Model section}
\label{fig:arch}
\end{center}
\end{figure}

As we have seen by the modules involved in the architecture, our proposed cognitive architecture is heavily inspired on the global workspace model \cite{dehaene1998neuronal}. The robot takes a decision based on several modules: long term and short term memory, an attentional system that filters information which is not relevant and its degree of happiness. The attentional system recieves a big set of information coming from the sensor of the agent in an analogy of the information coming from the senses that enter our brain. Despite the big amount of information that the senses perceive we are not conscious of all the information in each instant of time. We make our conscious decisions based on the information that we consciously perceive, which is a subset of the information processed by the brain. The unconscious mind filters the information that it considers to be useless for the conscious mind. Figure \ref{fig:arch} shows the structure of the cognitive architecture.
\section{Philosophy of mind connections}
Artificial consciousness can be seen as a sub-field from philosophy of the mind since the potential success of the implementations provided by the field depend on the assumptions made by functionalism, one philosophy of mind theory that is described on this section. Hence, it is important to review some philosophy of mind ideas and how the model proposed in this work is related  or could be a proper implementation of these ideas.

We will first consider Descartes dualism \cite{rozemond2009descartes}, where the mind is a separate entity than the body. In the dualism theory, the mind is a witness of the information that the senses coming from the physical world through the senses of the body give it. Let us define the information processing processes as System 1 and the conscious-self process as System 2 \cite{morewedge2010associative}. In our proposed system, we emulate the Descartes mind through the System 2 process, which emulates the conscious-self, that could emerge from another realm as Descartes states. The attentional, or unconscious emulator, system and the memories databases feed with information the conscious-self, being apart from others. Hence, if consciousness may arise from a system that process information as the human brain does and this process is independent from the rest, it could emerge in this process according to this theory. We could see the System 2 process as the viewer of a theatre \cite{baars1997theatre}, where the information is processed according to the global workspace model \cite{newman1997neural}.

If we consider, as in dualism in the extreme position of defining consciousness as a process that belong to a non-physical realm \cite{clarke2003descartes}, that the experiences that we feel are related with psychical, or non-physical properties that just correlate with brain processes, we can define these properties as qualia \cite{jackson1982epiphenomenal}. 

On the other hand, we find the physicalism theory, that basically defines that the mind states are a direct physical representation of our mental states, or the experiences that we feel \cite{nagel1965physicalism}. If mental states are not only correlated but completely explained by physical states we would be assuming that the identity theory is true \cite{smart2009identity}. Hence, if we could program those physical states in a machine, mental states may automatically arise in response to them. The states may be programmed, as in this proposal, as the level of an indicator and the weights of an adaptive system. These states activate in response to environmental stimuli and in doing so they causally impact on our behaviours \cite{kim2018philosophy}. We could argue that biological beings would only be the ones that develop consciousness but an objection resides in the fact biological beings are develop by DNA, which is essentially a code. If we are seen as machines \cite{searle1990brain} being developed by DNA \cite{dennett1996darwin}, why would not an agent as the one described in this work be able to develop consciousness as well?

But we could argue that consciousness may only arise in biological beings, due for example to chemical reactions and not in computers that implement Turing machines \cite{eliasmith2002myth}. If we assume that consciousness can emerge independently of the materials and reactions but dependently on the physical states and the functional processing of information, then, artificial consciousness could be implemented in computers. This is basically what functionalism defends \cite{lycan1974mental}. In functionalism, the mental state is defined by its position on a causal chain and its dependant on external estimuli, previous behaviour and mental states \cite{hierro2005filosofia}. The mental states are functional states, transformations performed over representations. In this case our robot has as mental states the ones given by its position, its memories, its happiness state and the weights given by the attentional system. Pain and other emotions could be simulated with other thresholds than can vary through the information recieved by sensors, emulating our senses. 

We could only assume that the proposed model or a generalization replacing all the adaptive systems based on weights by deep neural networks can make consciousness emerge if we agree with functionalist point of views. We can find in the literature which are the main objections to the functionalism theory \cite{block1978troubles,eliasmith2002myth}.

Other views on the mind include logical behaviorism, which basically states that the external behaviour of an agent explains its mental concepts \cite{kitchener1999logical}. That is, they are correlated. This thesis is the base of emulating consciousness in computers by simulating the correlated behaviours that humans perform when they are conscious \cite{merchan2020machine}. In this work the robot performs movements according to its memories and the level of the happiness indicator. We could predict how happy is the robot based on its decisions. This is an example of how mental states could be a mirror of external behaviours. The movements performed by the robot are also rational \cite{harman2013rationality} in the sense that they are a logical implication of internal and external outputs, as in conscious beings.
    
Other discussed theory for machine consciousness apart from the global workspace theory that could make machines conscious is the integrated information theory \cite{tononi2016integrated}. This theory is based on the following axioms of consciousness: It has intrinsic existence, my experience exists from my own intrinsic perspective; it has structure, it has to process a certain amount of information, it is integrated and excludes information. This amount of information is called the integrated information that the system must process in order to be conscious. The system would need to autonomously process cause and effect relations, which related to intrinsic perspective. These relations must affect the whole system to be structured, excluding non necessary information and being irreducible. Our proposed method is not based on this theory but in our robot the decisions taken have their cause on the happiness degree and the environment, having an intrinsic perspective. Then the robot makes a move based on that decision. We exclude from the decision taken by the System 2 process the information that the attentional or unconscious system qualifies as not important giving its weights. Following this reasoning, but excluding the idea of the minimum amount of information processed by an agent to be conscious, we also have influence from the integrated information theory.

It is arguable nonetheless that our agent performs behaviour with intentionality \cite{searle1983intentionality} or free will \cite{van1983essay}, as it is completely determined by the weights of the attentional system, the environment and the other modules. However, if we view this with a physicalist point of view \cite{jacob2003intentionality}, it would also be arguable that we perform our actions with free will as they would also be governed by the same things (unconscious filter of information entering from our senses, memories and levels of feelings as in the happiness case of our agent). Although, it seems to us that we can still have the freedom to decide \cite{jaworski2011philosophy}, which is the action that we model with the System 2 process in our agent. So, as we cannot directly measure consciousness in a system, it is hard to determine whether an agent has free will or intentionality from our point of view. It is important to distinguish consciousness from intelligence, so the Turing test \cite{saygin2000turing} or other measures of intelligence such as the Winegrad schema \cite{levesque2012winograd} are not related to determining whether an agent is conscious of itself.

Possible enhancements of our agent reside in being able to process abstract complex data representations as in the consciousness prior architecture \cite{bengio2017consciousness}. In this architecture, an attention neural network \cite{vaswani2017attention} sums up the information that the sensors of a robot perceive into a complex abstract representation of it that is processed by another deep neural network which entries are the output of the attention neural network and information coming from memories and the representations of the network at the previous instant of time. We agree in modelling cognitive architectures following these principles and it would be interesting that our robot would also learn these abstract representations. Another example is a machine consciousness architecture using deep learning and Gaussian process to emulate behaviors seen in humans when they are conscious \cite{merchan2020machine}. 
\section{Experiments}
In this section, we provide a big set of experiments to provide empirical evidence of the advantages that cognitive architectures such as the one described in this work could give to agents.

\begin{figure}[htbp]
\centering{
\includegraphics[width=5cm]{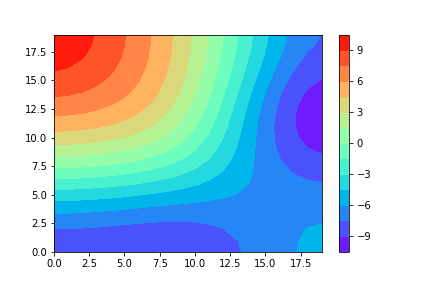}
\includegraphics[width=5cm]{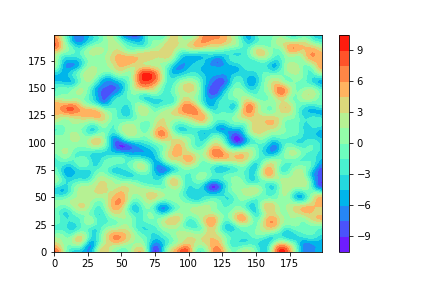}
}
\caption{Heat map of the first magnitude values of the environment ES20M5 and ES200M5.}
\label{small_env_example}
\end{figure}

We first define the environment that our model will interact with is represented as a list of $m$ magnitudes each of one can be represented as a bi-dimensional grid of size $N \times N$. Also, one condition that was established is that there should not be high differences in the magnitude value of any pair of adjacent cells. Each position of the grid holds several values $m$ that every sensor of the robot measures. A tri-dimensional grid is instantiated with size $N \times N \times m$. Each cell of the environment grid is initiated with a random real number, which is calculated using a normal distribution $\mathcal{N}(0,1)$. For each magnitude, all the values assigned to its bi-dimensional grid are normalized so their values lay within the interval $[-\alpha, \alpha]$. We generate multiple random environments with a different size of the grid $N=[20,100,200]$, whose first magnitude can be seen on Figure \ref{small_env_example} and number of magnitudes $m=[5,20,100]$. These environments are identified by ESxMy. Where $x$ is the grid size and $y$ the number of magnitudes.
In order to interpret the magnitude values into preferences, it is necessary to define the preference functions related to each magnitude. All of the defined functions satisfy that any magnitude value within $[-10, 10]$, returns a preference value within $[0, 10]$, with a preference of $0$ being the worst possible result for the agent and viceversa. The analytic expression of these functions are $ f(x) = |10 \cdot \sin{\left(\dfrac{x}{10} \cdot \dfrac{\pi}{2}\right)}|$, $f(x) = |10 \cdot \cos {\left(\dfrac{x}{10} \cdot \dfrac{\pi}{2} \right)}|$, $f(x) = 5 + 5 \cdot  \sin{\left(\dfrac{x}{10} \cdot \dfrac{\pi}{2} \right)}$ and $f(x) = -\cos{\left(\dfrac{x - 1}{10} \cdot \dfrac{\pi}{2} \right)}$.

The absolute sine function, gives a higher preference to extreme values, while giving a lower preference to magnitude values close to zero. The absolute cosine function has the opposite behaviour to the absolute sine function. This function is good for mapping most of the physical magnitudes that can be found in real life, for example temperature, since extreme values are, by definition, quite extreme and bad for both life and computational agents. The sine function, returns a higher preference for higher magnitude values and the additive inverse sine function, which acts as the opposite of the sine function. The preference functions are mapped to magnitudes randomly.

To be able to determine how good our model is performing, it is necessary to first define the evaluation metrics that are going to be used: FH: Final happiness. MH: Mean happiness. HSD: Happiness standard deviation NI: Number of iterations. WGFP: Weighted goodness of final position. MWG: Mean weighted goodness. MTI: Mean time per iteration (milliseconds). The final happiness is able to tells us how good was the agent at the moment of reaching the end of its path. The mean and standard deviation happiness are calculated with all happiness values obtained at each iteration of the simulation. Another goal of the model is to design agents that are able to localize and travel to the best possible position inside its environment, which is defined by the position that returns the best values of goodness for all of the magnitudes of the environment, using their corresponding preference functions. The global goodness of any position will be equal to the mean of all goodness values for each different magnitude. The sum of weights assigned to all four observations for one magnitude, one for each possible action, is considered the importance that is given to such magnitude at any specific iteration. Therefore, this value will be used as a weight for that magnitude at the moment of calculating the weighted goodness of the position of the agent.
The weighted goodness of the final positions is a good metric to evaluate how effective was the agent navigating into a position that satisfy its own idea of what is a good position.

The presented model has multiple variables that need to be defined for each simulation. Each parameter is tuned individually in all the environments according to several metrics. The initial parameters values considered are shown bellow: Number of iterations = $100$. Initial degree of happiness = $5$. Fatigue = $0.1$. Death threshold $\tau_{d} = 2$. Critical near death threshold $\tau_{c} = 3$. Risk threshold $\tau_{r} = 4$. Attentional conscious limit = $2$. Short-term memory capacity positions = $3$. Short-term memory threshold = $2$. Long-term memory capacity = $2$. Long-term memory threshold = $4$. Adaptation weights = $[0.05, 0.1, 0.2]$.

After initializing all variables with the values described, the model was tested against all the nine environments defined before. The obtained results can be observed at the table \ref{initial_param_eval}.
\begin{table}[htbp]
\small
    \centering
\begin{tabular}{l|rr|rrrrr}
\toprule
       ENV &     MH &    MWG &      FH &    HSD &   NI &   WGFP &    MTI \\
\midrule
    ES20M5 & 8.3425 & 3.2180 &  8.0033 & 2.0070 &  100 & 2.4248 & 0.3700 \\
   ES20M20 & 8.5641 & 2.3737 & 10.0000 & 2.0624 &  100 & 2.6874 & 0.6400 \\
  ES20M100 & 7.8666 & 3.3172 &  6.3860 & 1.7848 &  100 & 3.6951 & 2.3700 \\
   ES100M5 & 3.4002 & 6.5684 &  1.9108 & 1.0258 &   18 & 6.3144 & 0.3333 \\
  ES100M20 & 4.9274 & 3.6557 &  6.8798 & 0.9812 &  100 & 2.8410 & 0.7500 \\
 ES100M100 & 7.9255 & 2.0716 &  9.6474 & 2.4830 &  100 & 1.1165 & 2.7100 \\
   ES200M5 & 8.4133 & 1.6965 & 10.0000 & 1.8782 &  100 & 0.0932 & 0.4500 \\
  ES200M20 & 8.8552 & 1.6618 & 10.0000 & 1.9101 &  100 & 0.4386 & 1.4000 \\
 ES200M100 & 8.5203 & 2.6202 & 10.0000 & 2.0038 &  100 & 2.8214 & 2.5200 \\
 \midrule
 MEAN & 7.4239 & 3.0203 & 8.0919 & 1.7929 & 91 & 2.4924 & 1.1125\\
\bottomrule
\end{tabular}
    \caption{Result of evaluating the model with the initial set of parameters}
    \label{initial_param_eval}
\end{table}
There are multiple conclusions that can be made after analysing these results. First, it can be seen that out of all nine environments that were used to test the agent, it was able to survive in eight of them, only dropping its happiness under the death threshold in one environment, ES100M5. The over all mean happiness is also pretty high.

\begin{figure}[htbp]
  \centering
  \fontsize{8}{10}\selectfont
  \includegraphics[width=150pt]{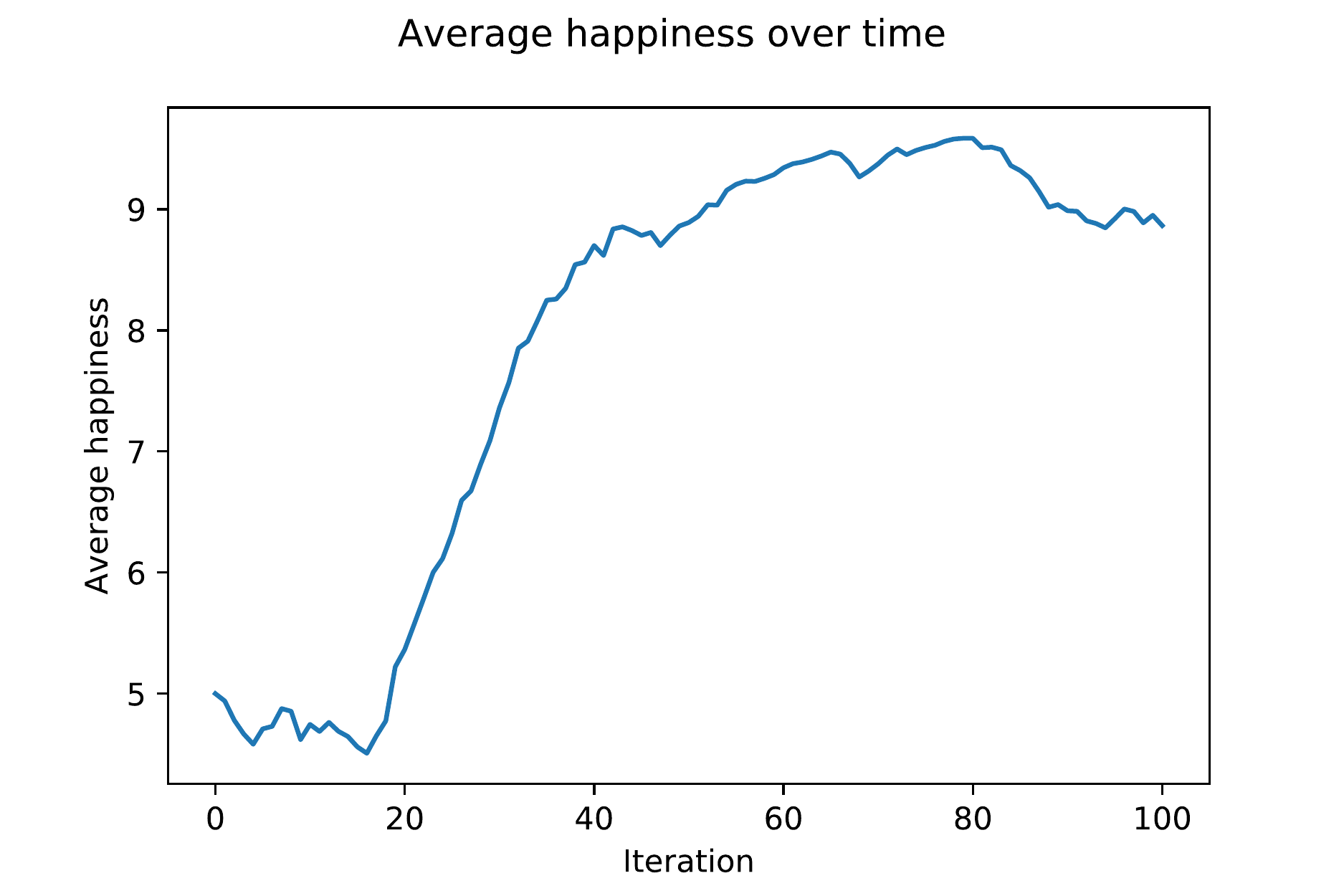}
  \includegraphics[width=150pt]{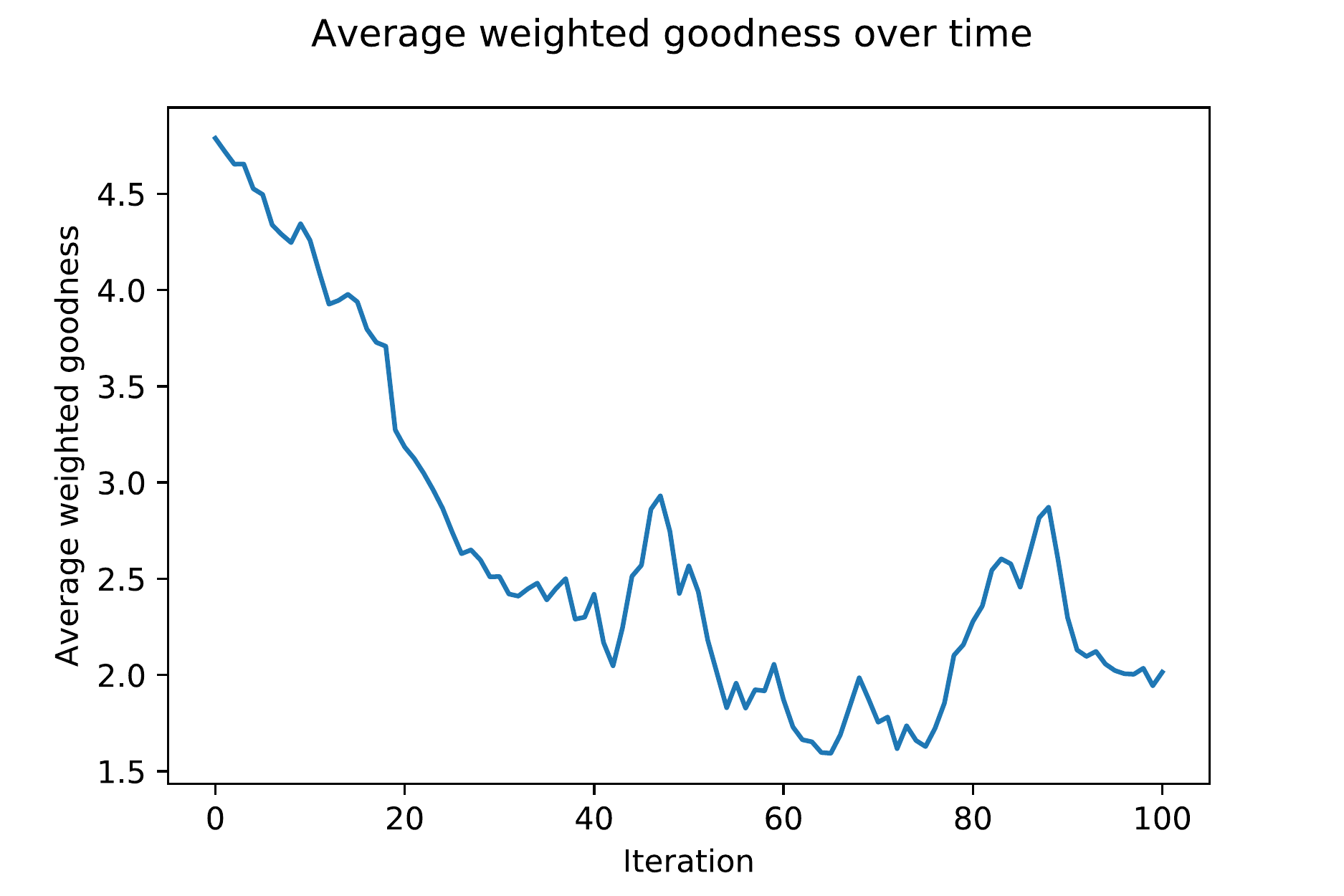}
  \caption{Average happiness (left) and weighted goodness (right) over a hundred iterations for the initial set of paremeters tested against the nine different environments}
  \label{avg_happiness_initial}
\end{figure}

The weighted goodness values of final positions can also be considered good. We can see that the highest MWG was the one obtained in the only simulation that ended with the agent death, which is the simulation performed in the environment ES100M5, with a MWG value of $6.5684$. Over all, all values of MWG are under $4$ for the eight surviving agents, and in two simulations the agent was able to achieve an WGFP under $2.0$, which is a very good result. There is a clear relation between the MTI and the number of magnitudes. This relationship is caused by the additional effort that has to be made by the attentional system in order to select the observations that will be passed to the global workspace. Even better results might be obtained after the calibration of all parameters of the model. In Figure \ref{avg_happiness_initial} is represented the average value of happiness over time for one hundred iterations in all environments.
It can be observed that the agent, in average, maintained its initial happiness, $5$, for around twenty iterations before its happiness started raising rapidly, up to the iteration sixty, when the agent maintained a constant value near $9$. Also, in Figure \ref{avg_happiness_initial} we can see the average weighted goodness over time. 

\begin{figure}[htbp]
  \centering
  \fontsize{8}{10}\selectfont
  \includegraphics[width=150pt]{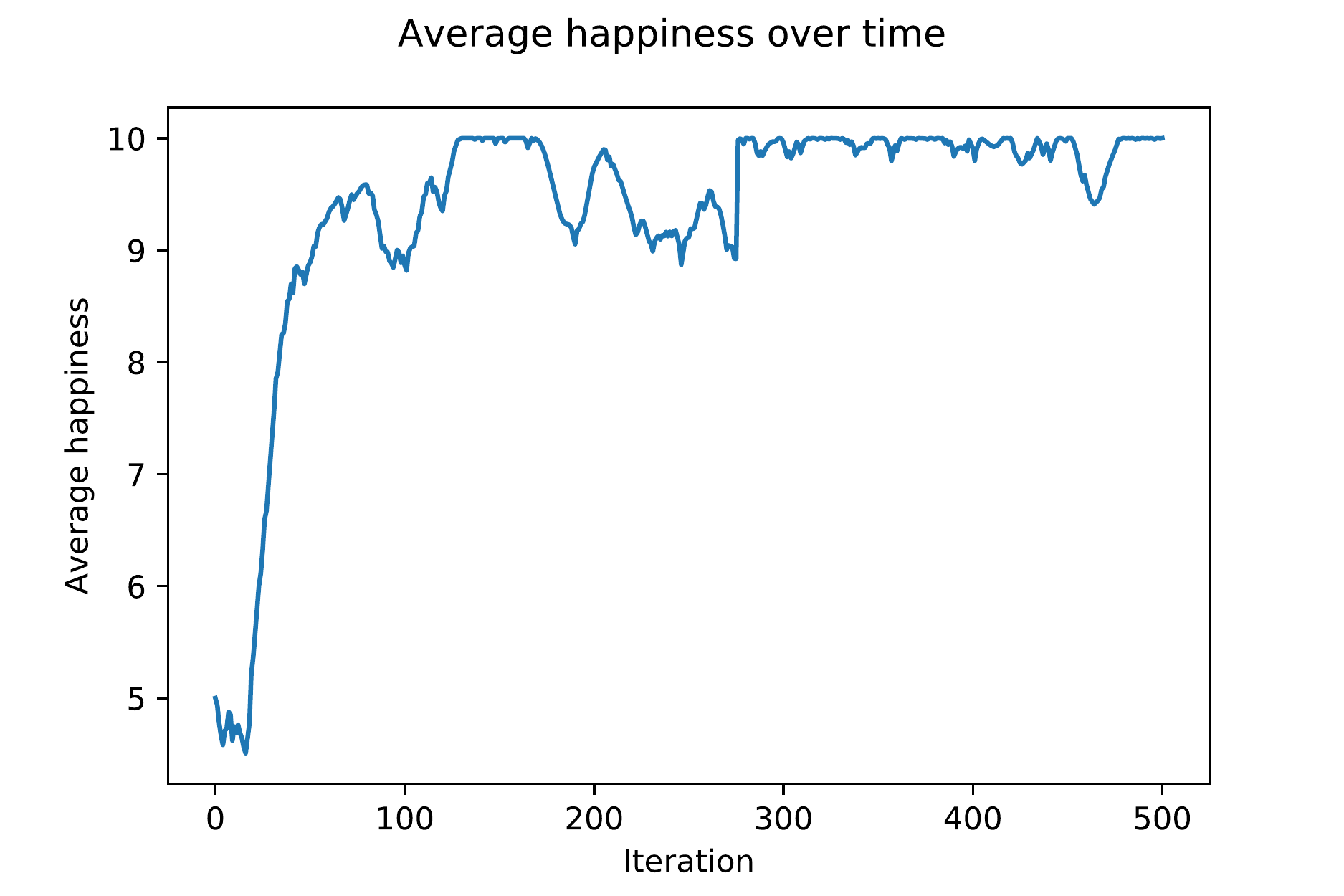}
  \includegraphics[width=150pt]{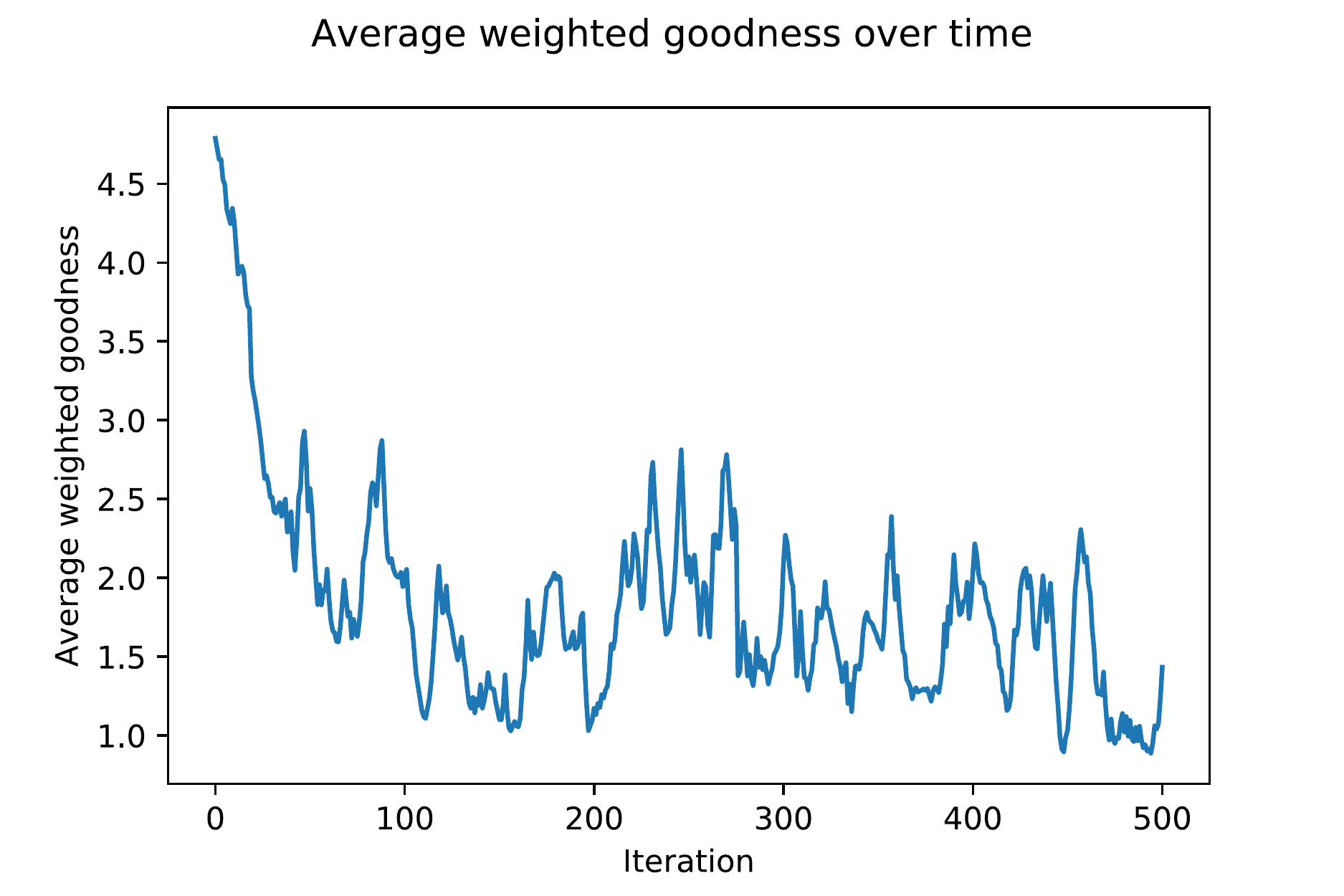}
  \caption{Average happiness (left) and weighted goodness (right) over a five hundred iterations tested against the nine different environments}
  \label{avg_happiness_500_iter}
\end{figure}

We can see a constant improvement of the average weighted goodness, which starts at a initial value of near $5$, and progresses over time up to an average of near $2$. There are some up and downs, but over all the graph shows that the progression follows a linear improvement over time. Given this progression, it is interesting to see if a lower value of weighted goodness could be achieved with a higher number of iterations. We repeat this simulation with a maximum number of iteration set to $500$, obtaining a similar set of results as in the case of $100$ iterations. We observe that one more agent died before reaching the 500 iterations, the one related to environment ES100M20, at the iteration number $275$. We plot the average happiness and weighted goodness in Figure \ref{avg_happiness_500_iter}.
It can be seen that there is a fast improvement during the first $100$ iterations and after this, however, agents seem to maintain on average a weighted goodness value between $1.0$ and $3.0$, and a happiness value between $9$ and $10$, converging. An interesting fact is that the direct relationship within happiness and the weighted goodness: when the latter goes down, so does the former. This can be observed in the section between the iterations $200$ and $300$. Therefore, the maximum number of iterations was set to $500$ for all the following simulations.

We now tune four happiness-related parameters exists in the model: fatigue, initial degree of happiness, and the two of the three happiness thresholds: critical near death threshold $\tau_{c}$, and risk threshold $\tau_{r}$.

The first happiness-related parameter to be tuned was the fatigue, which is a constant number that is substracted to the happiness of the agent in each iteration. We do not include more numerical results in tables for these parameters but talk about the conclusions extracted from of the experiments. In terms of mean happiness, a tendency to obtain higher happiness with lower values of fatigue can be noticed. In terms of mean weighted goodness, lower fatigue returns better results for all values. Still, we do not want a fatigue value that is too low, since we still want to punish agents that are performing bad for a long period of time. Therefore, it was decided for this analysis that the best value of fatigue was $0.05$.

The next parameter that was analyzed was the initial degree of happiness. As it was expected, a higher initial of happiness results in a higher mean happiness. The third happiness-related parameter that was tested was the risk threshold, $\tau_r$. There is not a noticeable change in performance when comparing the first two tested values, $4.0$ and $4.5$. The value for the risk threshold that was chosen due to resulting in a better performance was $4.5$.

Finally, the last happiness-related parameter that is tested is the critical near death threshold, $\tau_{c}$. The results for $\tau_{c} = 3$ and $\tau_{c} = 3.5$ does not have a noticeable difference. Using MWG as the decisive metric, the chosen for the critical near death threshold is $\tau_{c} = 3.5$. The attentional limit also takes an important role in the behaviour of the agent. It represents the number of magnitudes that is chosen by the attentional system in each iteration, which are later sent to the global workspace at the moment of taking a decision about which action to execute. Multiple simulations were executed testing five different values for the attentional limit, whose results can be seen in the table \ref{tuning_attentional_limit}.
\begin{table}[htbp]
\small
\centering
\begin{tabular}{r|rr|rrrrr}
\toprule
 attentional limit &     MH &    MWG &     FH &    HSD &       NI &   WGFP &    MTI \\
\midrule
1 & 7.9394 & 2.9516 & 7.2807 & 1.3820 & 373.0000 & 3.1601 & 1.0873 \\
2 & 8.7627 & 2.2434 & 8.1974 & 1.2392 & 431.4444 & 1.8643 & 1.1254 \\
3 & 8.6832 & 2.9691 & 8.0908 & 1.1957 & 425.5556 & 3.6208 & 1.1141 \\
4 & 8.4873 & 2.6407 & 7.4097 & 1.3014 & 444.2222 & 3.1423 & 1.1427 \\
5 & 8.8075 & 2.5964 & 8.1871 & 1.1160 & 435.3333 & 3.4292 & 1.1777 \\
\bottomrule
\end{tabular}
\caption{Mean results for different attentional limit values}
\label{tuning_attentional_limit}
\end{table}
The attentional limit $= 2$ has the best performance. Its value of mean weighted goodness, $2.2434$, is the lowest out of all, and its mean happiness, $8.7627$, is also very high, only improved by the mean happiness obtained with attentional limit $= 5$. We set the attentional limit to $2$.

Each of the memories of the agent has two parameters to be tuned, their capacity and the threshold. The first parameter that is analyzed is the short-term memory capacity. It can be seen that, MWG-wise, the best result is obtained with a capacity of five beliefs. We then proceeded with the other parameter of the short-term memory: its threshold, $\tau_{stm}$. In this case, it seems that decreasing the initial parameter from $\tau_{stm} = 2$ to $\tau_{stm} = 1.5$ produces slightly better results in terms of both mean happiness and mean weighted goodness, so the change is made permanent. We perform in an analogous way with the long-term memory. Here, it seems like there is a decent improvement of the model performance when increasing the long-term memory capacity from $2$ to $6$, going from a mean weighted goodness of $2.1758$ to $2.1025$. Regarding the long-term memory threshold, it seems like changing the initial values decrease the performance of the model. Therefore, the initial value of $\tau_{ltm} = 4$ stays intact.

Finally, the only parameters left to tune are the learning rates: the low learning rate, $\lambda_{l}$, the medium learning rate, $\lambda_{m}$ and the high learning rate, $\lambda_{h}$. Regarding the low learning rate, $\lambda_{l}$. It can be seen that changing the learning rate produces the higher difference in performance out of all the tested parameters, with a deviation of mean weighted goodness of almost $1$ when comparing some of the simulations. The best performance is found when the low learning rate is set to its initial value, $0.05$. 

We set $1.5$ for the final value of the medium learning rate. Lastly, regarding the high learning rate we can appreciate that the lowest value of MWG is obtained with $\lambda_{h} = 0.4$, but as we want the agent to be alive in critical situations, we set it to $2$.

\begin{figure}[htbp]
  \centering
  \fontsize{8}{10}\selectfont
  \includegraphics[width=150pt]{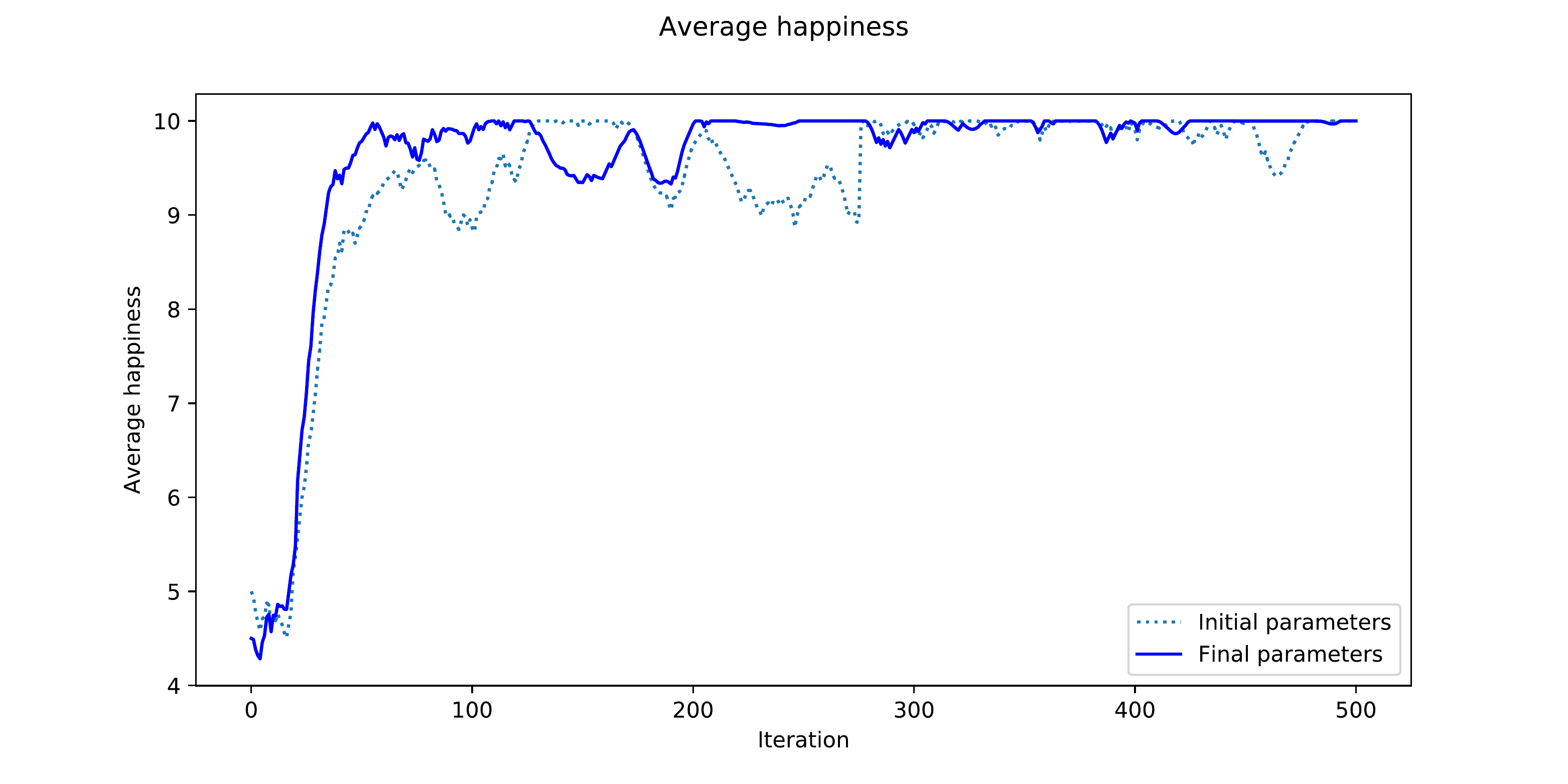}
  \includegraphics[width=150pt]{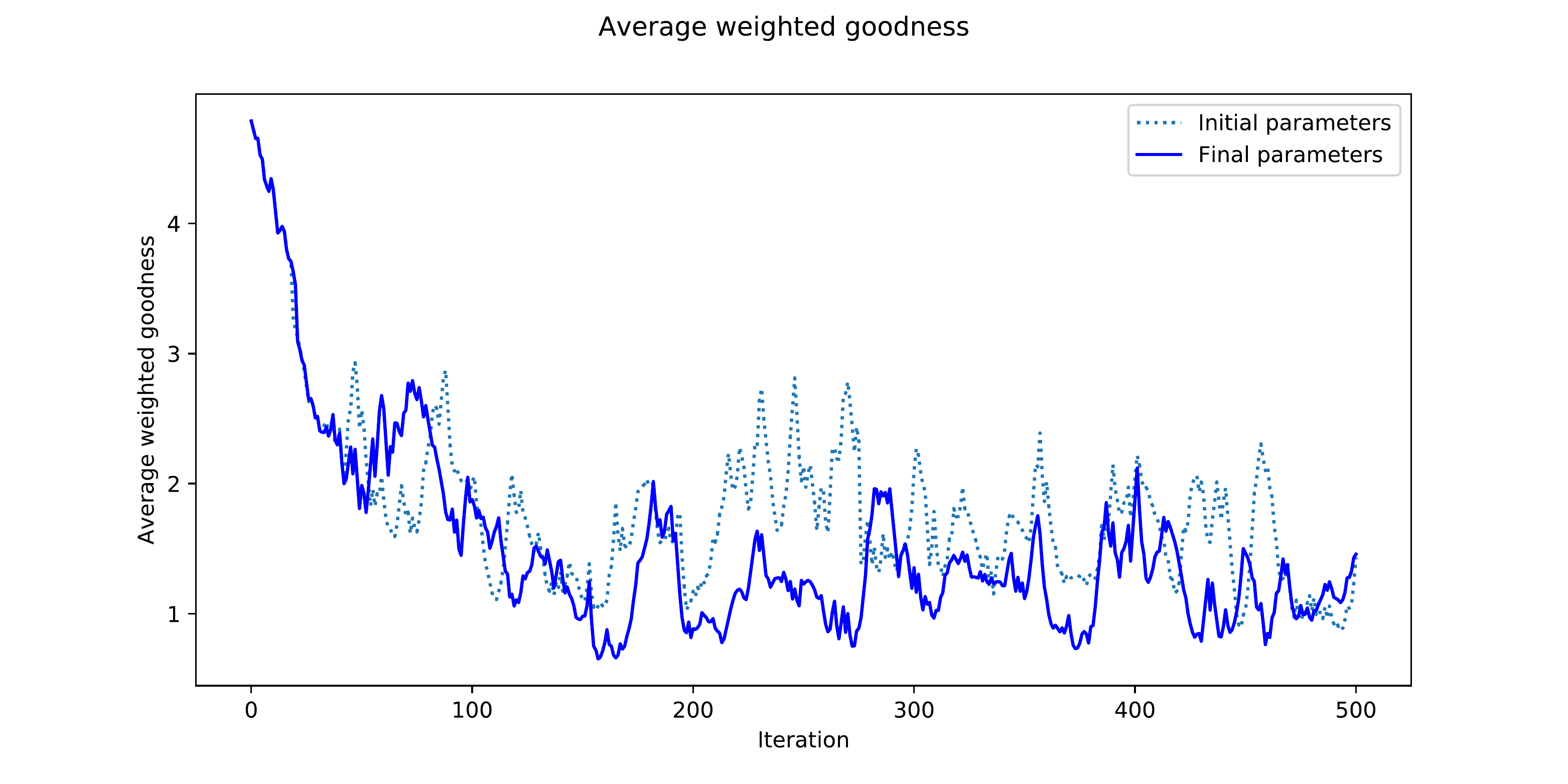}
  \caption{Average happiness (left) and weighted goodness (right) over a five hundred iterations comparing the final (solid line) and initial parameters (dotted line)}
  \label{avg_happiness_final}
\end{figure}

The results obtained for each of the nine tested environments, using the tuned set of parameters, are displayed in Table \ref{results_final_parameters}.
\begin{table}[htbp]
\small
\centering
\begin{tabular}{l|rr|rrrrr}
\toprule
       ENV &     MH &    MWG &      FH &    HSD &   NI &   WGFP &    MTI \\
\midrule
    ES20M5 & 9.6633 & 0.7233 & 10.0000 & 1.0984 &  500 & 0.0396 & 0.3760 \\
   ES20M20 & 9.4097 & 2.4639 & 10.0000 & 1.4519 &  500 & 1.8189 & 0.5760 \\
  ES20M100 & 9.6779 & 1.1043 & 10.0000 & 1.0570 &  500 & 0.0292 & 2.0100 \\
   ES100M5 & 3.2294 & 6.5653 &  1.9887 & 0.8067 &   20 & 6.5903 & 0.3000 \\
  ES100M20 & 9.5266 & 1.6654 & 10.0000 & 1.4315 &  500 & 3.0864 & 0.7060 \\
 ES100M100 & 9.5062 & 2.3734 & 10.0000 & 1.3311 &  500 & 1.1592 & 2.0680 \\
   ES200M5 & 9.5654 & 2.1843 & 10.0000 & 1.2407 &  500 & 2.2895 & 0.3300 \\
  ES200M20 & 9.7824 & 0.6477 & 10.0000 & 0.9607 &  500 & 2.6468 & 0.6760 \\
 ES200M100 & 9.7408 & 0.8837 & 10.0000 & 1.0270 &  500 & 0.5856 & 2.6000 \\
\bottomrule
\end{tabular}
\caption{Results of evaluating the model with the final set of parameters}
\label{results_final_parameters}
\end{table}
In order to analyze how well the final parameters perform compared to the initial set, both the average happiness and average weighted goodness over time have been plotted in Figure \ref{avg_happiness_final}.
On average, the final set outperforms the initial en both metrics. We can observe that peak performance is achieved near the iteration $150$, when an average weighted goodness near $0$ is achieved. The average happiness also shows good performance, keeping a value of near 10 at the iteration $100$, and maintaining such value for latter iterations.

Recall that any value of goodness lower than 5 is considered good. In this experiment, agents have shown to be able to maintain in average goodness values between $0$ and $2$, which by the definition given, are very good values. Therefore, we can deduct that the model is well constructed and the agents are achieving the goal that were set for them.

\begin{figure}[htbp]
  \centering
  \fontsize{8}{10}\selectfont
  \includegraphics[width=150pt]{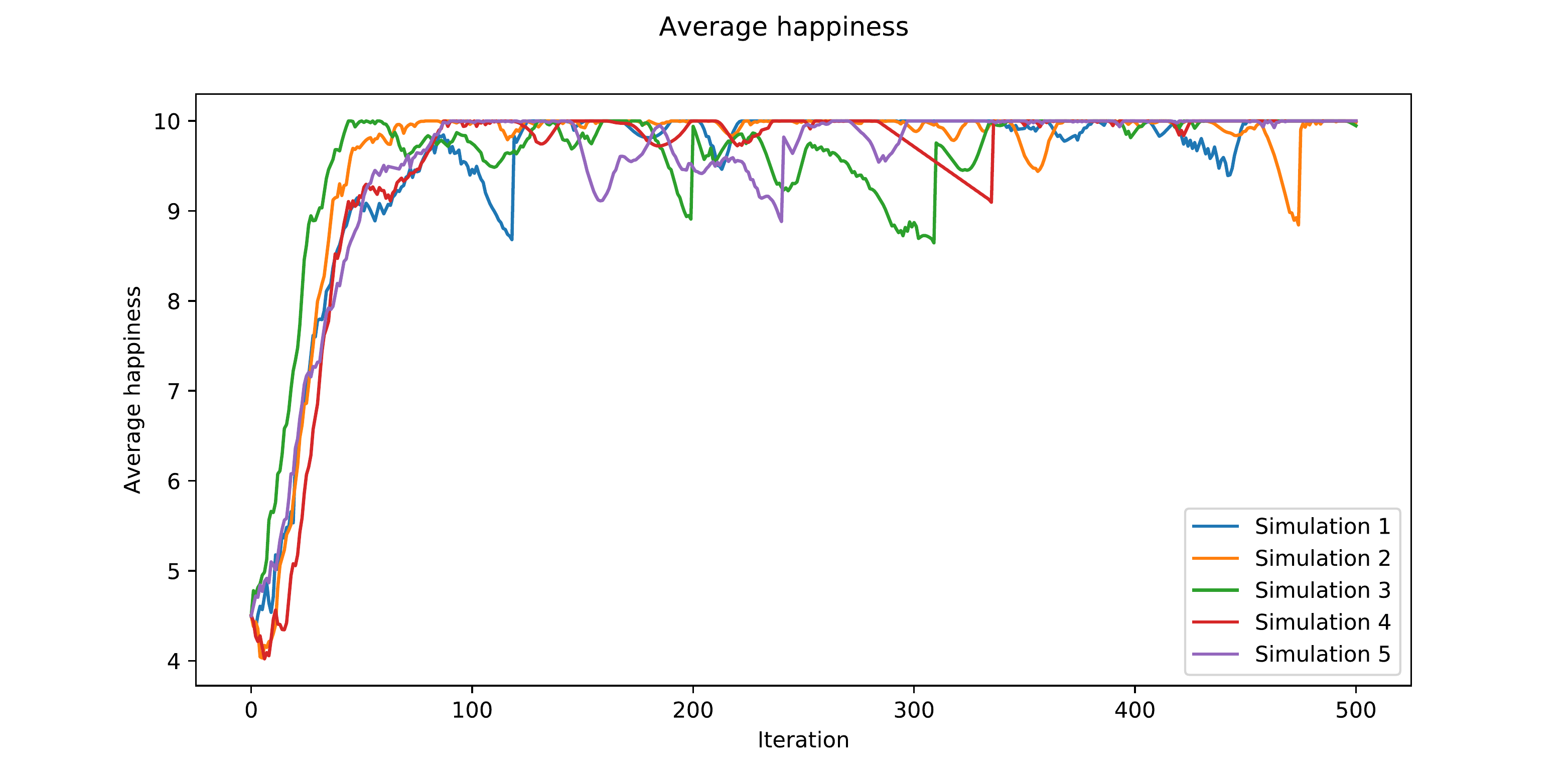}
  \includegraphics[width=150pt]{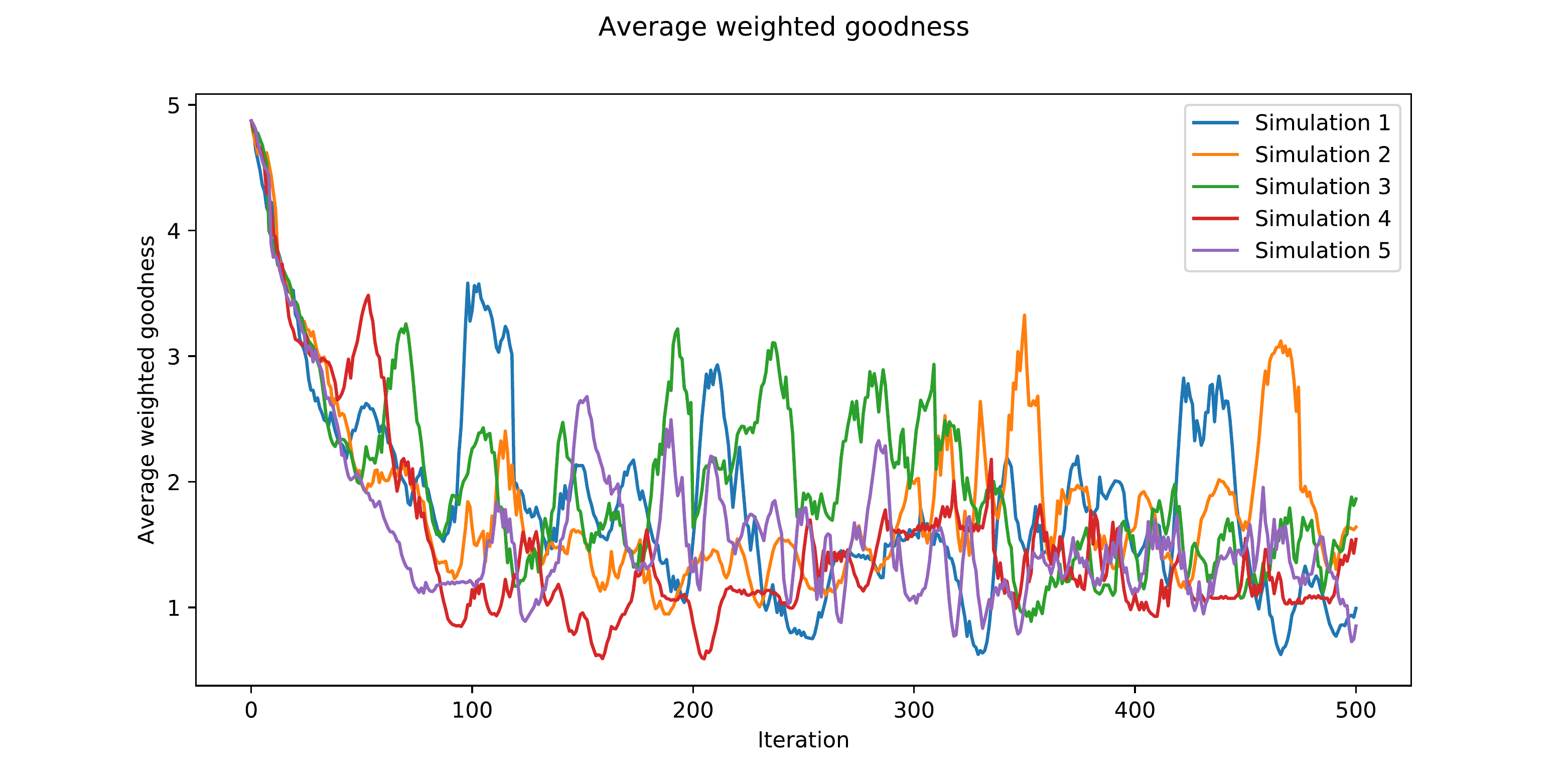}
  \caption{Average happiness and weighted goodness of the final model over five hundred iterations, in 5 different simulations}
  \label{avg_happiness_eval}
\end{figure}

We perform more experiments in five different simulations. In each simulation, nine new environments are randomly generated and each magnitude of each environment is assigned one of the four defined preference functions. The average results per each simulation can be found on the table \ref{results_eval}. 
\begin{table}[htbp]
\small
\centering
\begin{tabular}{r|rr|rrrrr}
\toprule
 Simulation &     MH &    MWG &     FH &    HSD &       NI &   WGFP &    MTI \\
\midrule
1 & 7.9834 & 2.8295 & 7.2776 & 1.2497 & 349.6667 & 2.8194 & 1.1530 \\
2 & 8.9011 & 2.4198 & 8.1681 & 1.1666 & 442.7778 & 2.9534 & 1.1381 \\
3 & 8.7009 & 2.9118 & 7.2381 & 1.1820 & 390.5556 & 3.9969 & 1.0910 \\
4 & 9.4744 & 1.5601 & 9.1094 & 1.4773 & 481.6667 & 2.0276 & 1.1219 \\
5 & 8.7284 & 2.4389 & 8.2148 & 1.2845 & 416.4444 & 1.9841 & 1.1809 \\
\bottomrule
\end{tabular}
\caption{Evaluation of the final model in 5 different simulations, for each simulation showing the average result of 9 different environments.}
\label{results_eval}
\end{table}
The average happiness and average weighted goodness are also shown over time in the plots of Figure \ref{avg_happiness_eval}.
We observe that all the results are similar: they all have a starting learning curve in the first 100 iterations and then a constant performance is maintained. It can be seen that in none of the simulation the mean weighted goodness has a value higher than $3$ so we can determine that the model is having a good performance. This experiment proves the capability of the model to generalize to different environments. We conclude that the evaluation of the presented model achieves to maintain high values of happiness and low values of weighted goodness.
\section{Conclusions and further work}
Along this manuscript, a prototype of a potentially conscious model has been proposed which has shown to create autonomous agents able to navigate through environments. The results of executing multiple simulations with different sets of parameters and different environments allows us to determine that the presented model shows a proper behaviour, being capable to reach and maintain high values of happiness. We show how the possession of feelings have proven to be a powerful tool which grants the system multiple beneficial abilities. Feelings are a great representation of the conscious state of an agent, and could serve as an example of a self-model representation. Two main advantages are provided by such self-model. First, it gives the agents the capability to report its conscious state as any moment. Secondly, it gives the agents the opportunity to have an understanding of its inner state and act according it to adapt to multiple situations. Attention filters the information that is way too big to process as a whole, deciding which observations are relevant enough to become conscious and which do not. The memory system, used to store past experiences and beliefs, also reinforces the adaptation ability that makes the agents able to learn from previous mistakes and successes. Using two different memories gives a higher degree of depth to this system. Finally, the global workspace, where decisions are made, has the information processed by the rest of the modules is integrated. If consciousness could exist within this model, it would lay here. 

As further work, we can implement more complex environments where the robot would need to interact in complex ways with the environment to remain alive. We will also implement deep neural networks for all the adaptive systems and memories of the robots, expanding and implementing the model in the same fashion as in the consciousness prior. As we have seen, the model has several parameters but we do not have an analytic expression to optimize them and computing the mean results can be expensive. Bayesian optimization has been used with success in similar problems \cite{garrido2018suggesting,cordoba2018bayesian}. In particular, we could optimize several metrics simultaneously under the presence of constraints such as the robot not dying with a constrained multi-objective Bayesian optimization approach \cite{fernandez2020max,garrido2020parallel}. Finally, we will implement several levels of feelings in the robot, for example: hungry, social needs or need to dream. The robot actions will be a function of that feelings along with happiness in a multi-objective problem where the robot will need to keep the level of all its indicators above some threshold.
\bibliographystyle{acm}
\bibliography{notes}
\end{document}